\documentclass[letterpaper, 10 pt, conference]{ieeeconf}  

\IEEEoverridecommandlockouts                              

\usepackage{graphics} 
\usepackage{epsfig} 
\usepackage{mathptmx} 
\usepackage{times} 
\usepackage{amsmath} 
\usepackage{amssymb}  
\usepackage[ruled]{algorithm2e}
\usepackage{subfigure}
\usepackage{tabu}
\usepackage{booktabs}
\usepackage{multirow}
\usepackage{footnote}
\usepackage{threeparttable}
\usepackage{booktabs}
\usepackage[table]{xcolor}
\usepackage[letterpaper,top=60pt,bottom=43pt,left=48pt,right=48pt]{geometry}

\title{\LARGE \bf
	Federated Imitation Learning: A Privacy Considered Imitation Learning Framework for Cloud Robotic Systems with Heterogeneous Sensor Data
}

\author{Boyi Liu$^{{1},{4}}$, Lujia Wang$^{1}$, Ming Liu$^{2}$ and Cheng-Zhong Xu$^{3}$
	\thanks{*This paper was recommended for publication by Okamura, Allison upon
		evaluation of the Associate Editor and Reviewers' comments. This research is supported by the Shenzhen Science and Technology Innovation Commission (Grant Number JCYJ2017081853518789), the Guangdong Science and Technology Plan Guangdong-Hong Kong Cooperation Innovation Platform (Grant Number 2018B050502009) and the National Natural Science Foundation of China (Grant Number 61603376) awarded to Dr. Lujia Wang. The National Natural Science Foundation of China (Grant Number U1713211) ,the Shenzhen Science, Technology and Innovation Commission (Grant Number JCYJ20160428154842603) and Basic Research Project of Shanghai Science and Technology Commission (Grant No. 16JC1401200) awarded to Prof. Ming Liu. The work also inspired and supported by Chengzhong Xu from University of Macau.}
	\thanks{$^{1}$Boyi liu, Lujia Wang are with Cloud Computing Lab of Shenzhen Institutes of Advanced Technology, Chinese Academy of Sciences. {\tt\small liuboyi17@mails.ucas.edu.cn};
		{\tt\small lj.wang1@siat.ac.cn}}
	\thanks{$^{2}$Ming liu is with Department of ECE, Hong Kong University of Science and Technology. {\tt\small eelium@ust.hk}}
	\thanks{$^{4}$Boyi liu is also with the University of Chinese Academy of Sciences.}
	\thanks{$^{3}$Cheng-Zhong Xu is with the University of Macau.  {\tt\small czxu@um.edu.mo}}
}

\begin{document}

	\maketitle
	\thispagestyle{empty}
	\pagestyle{empty}

	\begin{abstract}
		Humans are capable of learning a new behavior by observing others perform the skill. Similarly, robots can also implement this by imitation learning. Furthermore, if with external guidance, humans can  master the new behavior more efficiently. So how can robots achieve this? To address the issue, we present Federated Imitation Learning (FIL) in the paper. Firstly, a knowledge fusion algorithm is proposed for the cloud fusing knowledge from local robots. Then, a knowledge transfer scheme is presented to facilitate local robots acquiring knowledge from the cloud. With FIL, a robot is capable of utilizing knowledge from other robots to increase its imitation learning in accuracy and  training efficiency. FIL considers information privacy and data heterogeneity when robots share knowledge. It is suitable to be deployed in cloud robotic systems. Finally, we conduct experiments of a simplified self-driving task for robots (cars). The experimental results demonstrate that FIL increases imitation learning efficiency and accuracy of local robots in cloud robotic systems.
	\end{abstract}

	\section{INTRODUCTION}
	In tradition imitation learning scenarios, demonstrations provide a descriptive medium for specifying robotic tasks. Prior work has shown that robots can acquire a range of complex skills through demonstration, such as table tennis [1], drawer opening [2], and multi-stage manipulation tasks [3]. Nevertheless, there exists a number of problems in the application of imitation learning. For example, large amounts of data are required and sometimes isomorphic. Particularly, the information of one robot cannot be shared with other robots. These drawbacks result in long training time for the robot and limited generalization performance. 
	\begin{figure}[thpb]
		\centering
		\includegraphics[width=0.9\linewidth]{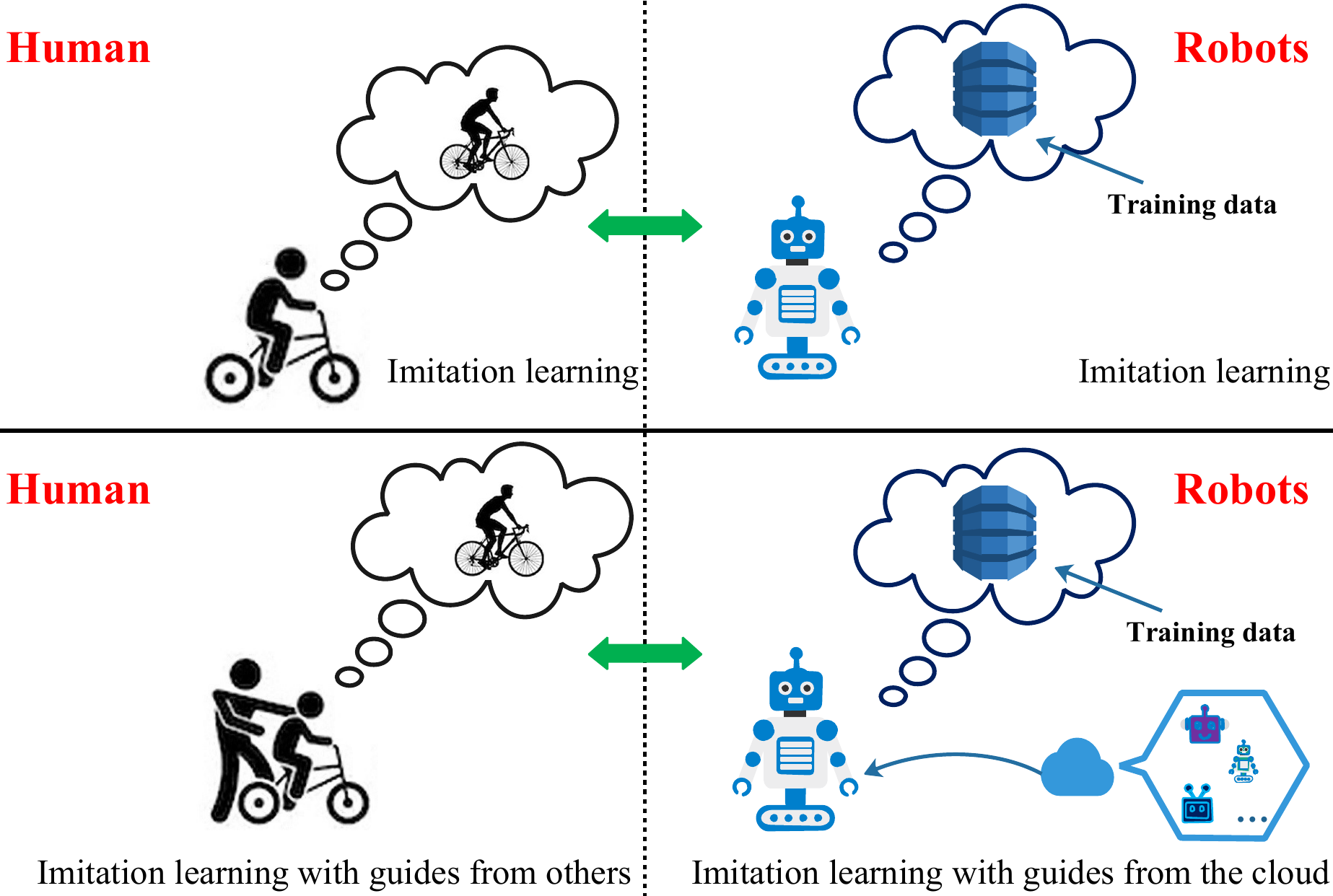}
		\caption{The child on the upper left acquires the ability to ride a bicycle by observing an adult. This is the process of imitation learning in humans. Correspondingly, the upper right robot acquires skills by training data. This is the process of imitation learning in robots. The bottom left child not only acquires bicycling skills by observing an adult, but also gets helps from an adult. This makes his learning more efficient. Inspired by this, in this work, FIL enables the bottom right robot not only acquires skills by training data, but also gets knowledge from other robots through the cloud robot system. So FIL increases imitation learning of local robots.}
		\label{fig:architecture}
	\end{figure} 
	\textbf{In the paper, we address the problem of how to improve imitation learning of robots in accuracy and efficiency by taking advantage of knowledge from other robots in cloud robotic systems. Moreover, we consider the premise of privacy protection and heterogeneous data.} 
	
	To realize the goal, we focus on cloud robotic systems [4]. Therefore, the Federated Imitation Learning (FIL) is proposed in the paper. As shown in Fig.1, It is inspired by the case that humans can learn more effectively  if they have external guidance in imitation learning. With FIL, a robot is capable of taking advantage of knowledge from other robots to increase its learning efficiency and accuracy. Even more,  without directly raw data sharing, FIL considers information privacy and data heterogeneity when robots share knowledge. So it is suitable to be deployed in cloud robot systems with heterogeneous sensor data. To evaluate it, we test FIL in an autonomous driving task. Experimental result indicates that FIL enables robots absorb knowledge from other robots and apply the knowledge to increase imitation learning efficiency and accuracy. Videos of the results can be found on the supplementary website\footnote[1]{The video is available at https://sites.google.com/view/federated-imitation}. Overall, this paper makes the following contributions:
	\begin{itemize}
		\item We present a novel framework named FIL. It increases imitation learning efficiency and accuracy of cloud robotic systems which may have difficulties in heterogeneous sensor data sharing and privacy protection consideration.
		\item We propose a knowledge fusion algorithm in FIL. It enables the cloud to fuse knowledge from other robots and generate guide models for robots with service requests.  
		\item Based on transfer learning, we present a knowledge transfer scheme to facilitate local robots acquiring knowledge from the cloud.
	\end{itemize}
	\section{Related Theory}
	Imitation learning is an effective way in various robotic task implementations. Besides, the efficiency and accuracy of imitation learning deployed on robots will be improved if with the cloud robotic system. Knowledge sharing and transferring are essential elements for cloud robotic systems and the shared knowledge can improve imitation learning in local robots. Typically, heterogeneity of robots or privacy protection will make it difficult for cloud robotic systems to realize knowledge sharing directly. For example, local robots are always have different types of sensors and they can't communicate with each other. In addition, there exists an inherent conflict: knowledge privacy protection and knowledge sharing. This conflict makes the cloud robotic system fall into the dilemma of information island. Nevertheless, federated learning is a key  method to solve this contradiction. With the idea of federated learning, the FIL realizes the sharing of the knowledge that is acquired through imitation learning. The cloud robotic system has the advantages of storage and computing, which can continuously evolve and expand the shared knowledge. So, we first briefly review the state-of-the-art works with respect to the cloud robotic systems, imitation learning and federated learning, among others. 
	\subsection{Imitation learning}
	Imitation learning is a problem in machine learning that autonomous agents face when attempting to learn tasks from another, more-expert agent. The expert provides demonstrations of task execution, from which the imitator attempts to mimic the expert’s behavior [5]. At present, imitation learning has been applied to a variety of tasks, including autonomous flight [6], articulated motion [7], and end-to-end driving [8, 9]. Imitation learning is mainly divided into two methods: behavioral cloning and inverse reinforcement learning. This work is based on behavioral cloning. In order to learn a skill, it is necessary to determine the most effective strategy representation. The most effective representation of a strategy is often the direct mapping of state features to trajectory actions, which is called behavioral cloning (BC). BC is a method of using supervised learning from state to action with given state-action teaching data set. For robotic systems, BC mainly focuses on behavior trajectory planning. System dynamics is not a key factor in it, so BC is a good choice.
	
	Early imitation learning studies interpreted the model-free BC approach as supervised learning. BC has been used in various applications. For instance, it has recently been used in the context of autonomous driving [10] and autonomous control of grasping [11]. BC is powerful when the agent requires only demonstration data to directly learn an imitation policy and does not require any further interaction between the agent and the environment. In [12], a neural network was trained in 1989 for an automated driving system. The work built a model from camera image to steering angle mapping. But it is not successful in practice. Reasons for failure include the constraints of the computing power and the size of the training set. However, [9,13] accomplished similar works and achieved success in a physical environment. These work has improved the teaching data set after using simulation platforms. At present, the BC approach can be rather brittle due to some well-known problems, such as limited data set of teaching, various state of apprentices' distribution of teaching data set. Some work has improved this issue, for example, one-shot imitation is achieved in [14]. However, it is still limited to a single robot and does not take system-level applications into account. Our work improved this issue, the teaching data set can be expanded infinitely with the cloud robotic framework. FIL compensates for distribution differences and increases training speed. FIL accomplishes it by enabling robot to acquire knowledge from other robots in cloud robotic system.
	
	\subsection{Cloud robotic system}
	Cloud robotic system is an emerging field in recent years. Cloud robotic system usually relies on many other resources from a network to support its operation. Since the concept of the cloud robot was proposed by Dr. Kuffner in 2010 [15], the research on cloud robots is rising gradually. Cloud robotics is an emerging field of robots embedded in networking, cloud computing, cloud storage [16]. While providing advantages of powerful computation, storage, and communication resources of modern data centers, it also allows robots to benefit from the platform which includes infrastructure and shared services [17]. Concretely, this approach has been used in mapping [18] and localization [19], perception [20], grasping [21], visuomotor control [22], speech processing [23], and other applications. Cloud robots have high requirements for communication. Wang et al. present a framework targeting near real-time MSDR, which grants asynchronous access to the cloud from the robots [24]. Sandeep Chinchali et al. present network offloading policies for cloud robotics [15]. It is worth noting that with the rapid development of artificial intelligence, knowledge sharing has become a prominent advantage of cloud robots. However, no specific privacy-considered knowledge sharing approach for cloud robots has been proposed up to now. We believe that the work present in this paper is the first privacy-considered knowledge sharing approach for cloud robotic systems with heterogeneous sensor data.
	
	\subsection{Federated learning}
	Federated learning was first proposed in [26], which showed its effectiveness through experiments on various datasets. In federated learning systems, the raw data is collected and stored at multiple edge nodes, and a machine learning model is trained from the distributed data without sending the raw data from the nodes to a central place [27, 28]. Federated learning is a machine learning method that is capable of protecting the privacy of users. It solves the problem of data silos. In practice, island data has different distributions. And it is reflected by different type of sensor data in cloud robotic systems. To address the issue, we present a knowledge fusing algorithm in FIL. FIL cleverly utilizes the simulation platform to achieve simultaneous acquisition of heterogeneous data. It enables federated learning to be applied to cloud robotic systems with heterogeneous data.
	
	Generally, this paper focuses on developing an imitation learning framework for cloud robotic systems, which is capable of protecting privacy and acclimatizing heterogeneous sensor data condition. This framework is well fit in cloud robot systems.
	\section{Methodology}
	The objects in the FIL framework include local robots, cloud servers, and requesting service robots. In FIL, local robots acquire skills through imitation learning. The cloud integrates the skills of the local robots and provides a corresponding cloud model. The cloud model serves as a guide model for the requesting service robot. Throughout the process, FIL considers the privacy of local robots so the cloud does not allow access to raw training data of local robots. FIL considers the heterogeneity of local robots' training data. It realizes the fusion of heterogeneous knowledge with the algorithm presented in the paper. FIL is capable of increasing imitation learning without sacrificing privacy of robots in cloud robotic systems. It enables robots to learn skills with Cloud-Robot-Environment setup. Communication devices and a cloud server are supporting facilities of FIL. Unlike distributed algorithms such as A3C or UNREAL, FIL can be executed synchronously or asynchronously. Because cloud servers in FIL only need to get network parameters rather than gradients, It avoids the confinement of communication in traditional cloud robotic systems. In summary, FIL has the following advantages:
	\begin{enumerate}
		\item FIL has the ability to fuse heterogeneous knowledge;
		\item FIL is capable of protecting the privacy of local robots;
		\item FIL can be executed synchronously or asynchronously, which reduces the requirement of communication.
	\end{enumerate}
	\subsection{Framework of FIL}
	The framework of FIL is performed in Cloud-Robot-Environment setup. There are local robots, cloud servers, communication services and computing device. Local robots learn skills through imitation learning and the cloud server fuses knowledge. We develop a federated learning algorithm to fuse private models into the shared model in the cloud. Then the cloud server is capable of generating guide models corresponding to requests of local robots. After that, the local robots perform transfer learning based on the guide model. Finally, the final policy will be quickly obtained. As illustrated in Fig.2, we explain the methodology in FIL with the example of a simplified self-driving task.
	\begin{figure}[thpb]
		\centering
		\includegraphics[width=0.85\linewidth]{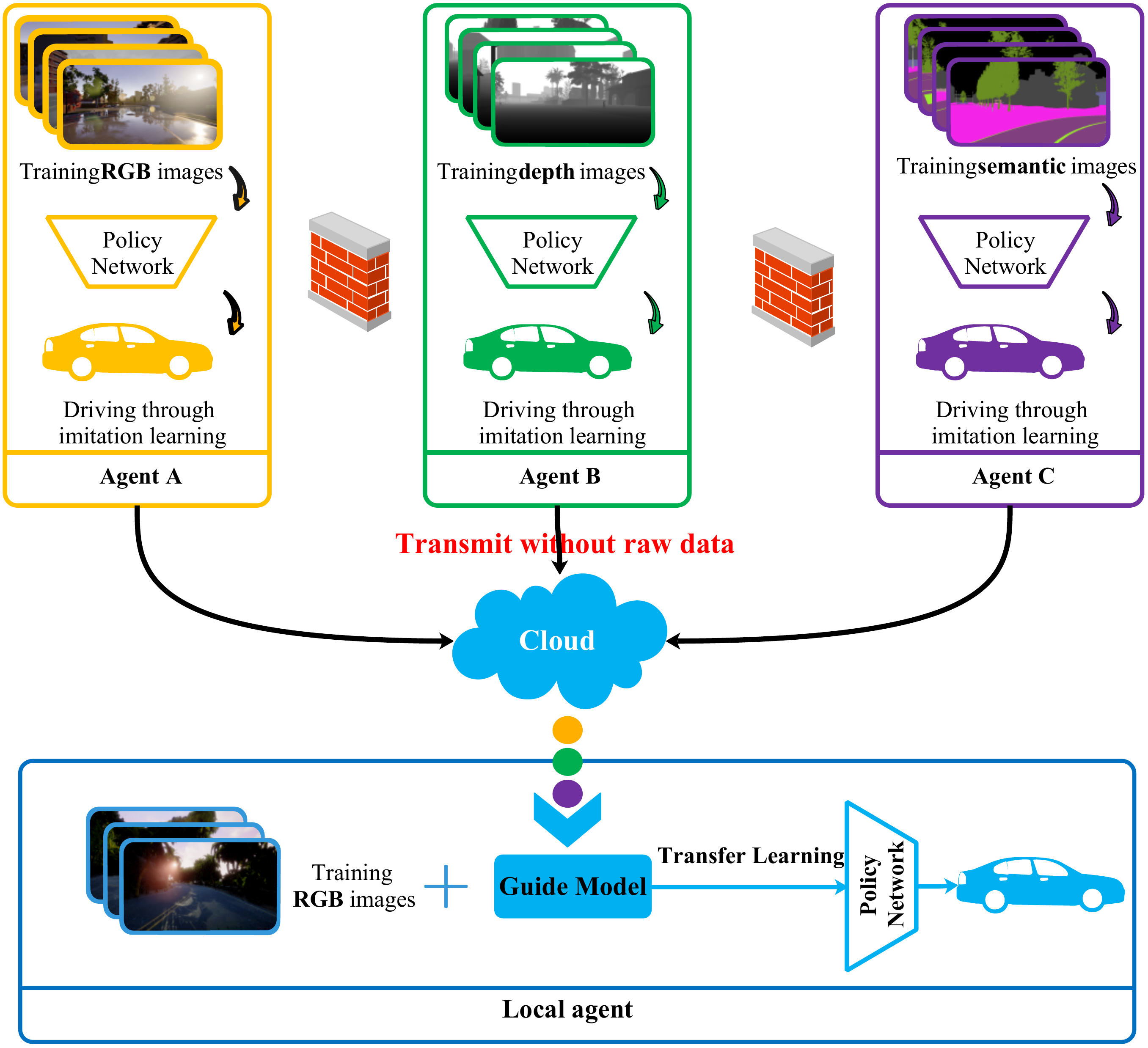}
		\caption{FIL framework. The work assume that there are three agents to fulfil the task. They perform imitation learning with heterogeneous data: RGB images, depth images and semantic segmentation images. Raw training data is not allowed to be shared between agents. Cloud does not have the right to obtain the original training data from agents either. Agents perform imitation learning to acquire the policy models. Only the parameters of the policy models are uploaded to the cloud. The cloud fuses the knowledge and then provide guide models to robots that request services. It facilitates the process of imitation learning.}
		\label{fig:architecture}
	\end{figure}
	
	In order to accomplish the simplified self-driving task, the three agents use three different types of training data separately. The training data is labeled and can be obtained by manually or simulation platform. Agent A trains RGB images and obtains a private policy model named private model A. The model takes RGB images as input. It outputs actions or the parameters that can determine actions. Similar processes occur in agent A and agent B. Outputs of the three private models are with same types. The inputs to the three models are different: RGB images, depth images, and semantic segmentation images. Then the parameters of the three models are uploaded to the cloud. The cloud fuses these models. Different from local data base, the cloud has its own data base. The data base in cloud has no labels. The cloud server labels the data with the uploaded private models. That is the process of knowledge fusion. Then it is capable of generating guide models with different types of input. When a local robot requests a service, the cloud provides a guide model corresponding to the type of sensor data. FIL can be performed online or offline. The framework of offline FIL can be summarized into four steps: Step1: Local robots perform imitation learning; Step2: Transmit parameters of private models; Step3: Fuse knowledge in the cloud; Step4: Provide guide models when there are requests from local robots.When online, step1 and step 2 are simultaneous. Labels of the cloud data will be updated simultaneously, as presented in Algorithm 1. 
	\begin{algorithm}
		\caption{Processing Algorithm in FIL}
		Initialize action-value Q-network with random weights ${\theta }$;
		
		\KwIn{$n$: number of local robots ; $f$: update frequency.} 
		\While{cloud server is running}
		{
			${\theta }_{nt}\leftarrow {robot }_{n}$ performs imitation learning
			
			\If{$t\%f==0$}
			{
				\For{$i=0;i<n;i++$} 
				{ 	
					Send ${\theta }_{it}$ to the cloud;
				} 
				labels=fuse(${\theta }_{1t}$, ${\theta }_{2t}$,$\cdots$,${\theta }_{nt}$)
			}
			\If{service\_request=True}
			{
				Generate ${\theta }_{cloud}$ base on labels;
				Send ${\theta }_{cloud}$ to local robots;
				
				${\theta }_{local}$=transfer(${\theta }_{cloud}$).
			}
		}
	\end{algorithm}
	\begin{figure*}[thpb]
		\centering
		\includegraphics[width=0.9\linewidth]{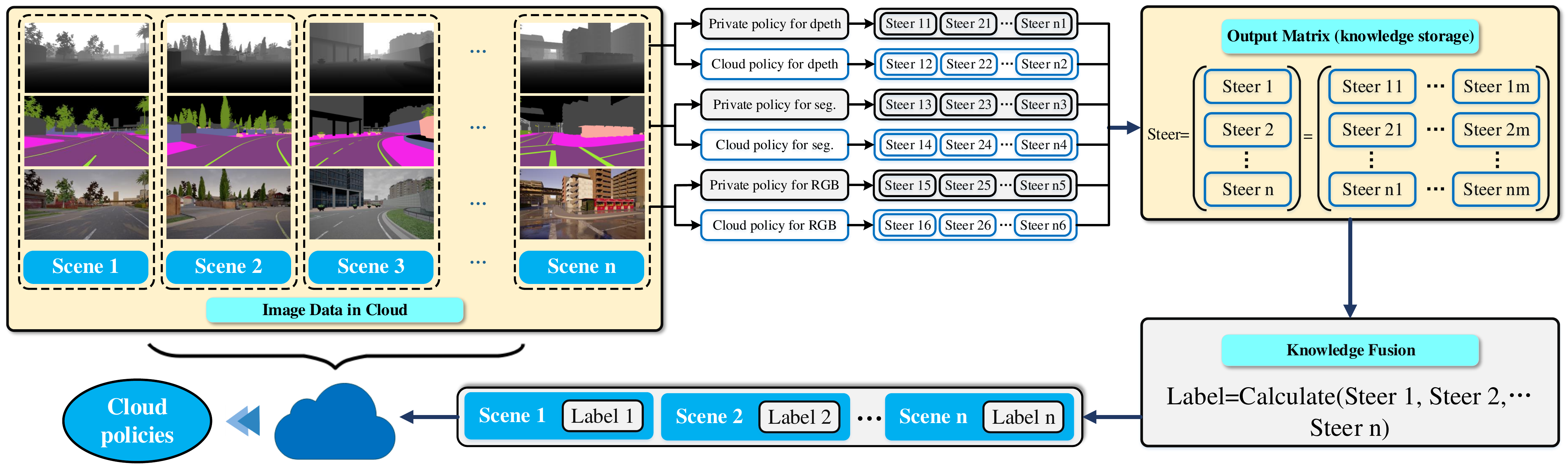}
		\caption{Knowledge fusion algorithm deployed on the cloud in a self-driving case. Agent obtain private models by performing imitation learning. The cloud stored different types of sensor data of every scenes. Input corresponding sensor data to private models.  Then calculate the numerical characteristics of private models outputs to label scenes. With multi types of sensor data, the cloud is capable of generating guide models corresponding to the sensor type of the local robot.}
		\label{fig:architecture}
	\end{figure*}
	
	Key algorithms in this framework includes the knowledge function algorithm and transferring approaches. We present them in the following.
	\subsection{Knowledge fusion algorithm in cloud}
	In traditional federated learning algorithms, heterogeneous data training is still difficult to achieve. But in the cloud robot system, we ingeniously utilize the function that the robot simulation system can collect different sensor data for a same scene. Base on this, we develop a knowledge fusion algorithm to obtain the shared model in cloud. The algorithm is efficient to fuse parameters of networks trained from different local robots that may have heterogeneous training data. The algorithm receives local network parameters and fuses them into sharing networks. 
	
	Fig.3 presents the knowledge fusion algorithm deployed in the cloud. The cloud server's primary responsibility is to label scenes. Before this, it is required to collect different types of sensor data that depends on the types of sensor data in local robots. For example, in the simplified self-driving task presented in the paper, the cloud needs to collect RGB images, depth images and semantic segmentation images. So one scene has three types of images in the cloud. Every type of images has a corresponding private model. Based on this, private models provide its own label suggestion of scenes to the cloud. The cloud will label these scenes with suggestions from the private models. The calculation approach of labels can draw on some methods of ensemble learning. It can be defined according to different application scenarios. For instance, in the self-driving case presented in the paper, we take the median of outputs. Because in this task, private models output the steer of the agent. Agents usually make decisions of along the road, turn, obstacle avoidance and so on. Generally speaking, the output of the model includes two extreme cases: turning and no turning. Errors in decisions often occur in the two extreme cases. Median can avoid extreme values in the evaluation, so we take the median of private models outputs as the label of the current scene. With data and labels, cloud models will be trained immediately. The process can be summarized to Formula (1) to Formula (5):
	\begin{equation}
	R_{{D_i}}^{emp}(\theta)=\frac{1}{N} \sum_{n=1}^{N}{L}\left({{y}_i}^{(n)}, f\left({{x}_i}^{(n)} ; \theta\right)\right)
	\end{equation}
	In the Formular (1), ${D_i}=\left\{\left(\mathbf{x}^{(n)}, y^{(n)}\right)\right\}_{n=1}^{N}$ is the data set of the local robot i. L represents the loss function.$\theta$ repensents parameters of models. $x_i$ are original data and $y_i$ are labels. N is the number of samples of the data set.
	\begin{equation} 
	\theta_{i}^{*} =\underset{\theta}{\arg \min } {R}_{{D}}^{struct}(\theta)
	\end{equation}
	The Formular (2) presents training targets of local robots. It is the structure risk minimization criteria.
	\begin{equation} 
	l_{in} =f_{{\theta}i}\left(scene_{in}\right)
	\end{equation}
	In the above formular, $scene_{in}$ is the training data in the cloud and i represents sensor types.
	\begin{equation} 
	Me_{in} =Median\left(l_{in}\right)
	\end{equation}
	\begin{equation} 
	\theta_{i(cloud)}^{*} =\underset{\theta}{\arg \min } \frac{1}{M} \sum_{n=1}^{M} {L}\left(Me_{in}, f\left(scene_{in}^{(n)} ; \theta\right)\right)+\frac{1}{2} \lambda\|\theta\|^{2}
	\end{equation}
	In Formular (4) and (5), $Me_{in}$ is the median of $l_{in}$, $\theta_{i(cloud)}^{*}$ is the training targets in the cloud. M represents the number of sample data. $\theta$ is the regularization term of L2 norm, which is used to reduce the parameter space and avoid over-fitting and lambda is used to control the intensity of regularization.
	
	It should be explained that the shared model in the cloud is not the final policy model of the local robot. We only use the shared model in the cloud as a guide model for local robots. The shared model maintained in the cloud is a cautious policy model but not the optimal for every robot. That is to say, the shared model in the cloud will not make serious mistakes in some private unstructured environments but the action is not the best. It is necessary for the robot to train its own policy model based on the shared model from the cloud. In order to reduce the error rate, we should transfer the shared model and train a new private model in local environments. This responded to the transfer learning process in FIL.
	\subsection{Transfer the shared model}
	There have been a lot of valuable studies on transfer learning. Current research mainly focuses on transferring through the relationship between source domain distribution and target domain distribution. This method is theoretically infeasible in cloud robotic systems. If local robots upload their training data to the cloud, the communication resources they need are unbearable. Therefore, cloud training data and local training data are not permitted to be calculated at the same time. Transfer learning from cloud to local belongs to supervised transfer learning. Moreover, the local robot can get the model parameters but without training data in the cloud.
	
	Under the constraints of the above conditions, Layer Transfer is the transferable learning method that can be implemented. Layer Transfer means that some layers in the model trained by source data are copied directly and the remaining layers are trained by target data. The advantage of this is that target data only needs fewer parameters to be trained, thus avoiding over-fitting. It has faster training speed and higher accuracy. On different tasks, the layers that need to be transferred are often different. For example, in speech recognition, we usually copy the last layers and retrain the first layers. This is because the first layers of speech recognition neural network are the way to recognize the speaker's pronunciation, and the last layers are the recognition. The latter layers have nothing to do with the speaker. In image recognition, we usually copy the front layers and retrain the back layers. This is because the first layers of the image recognition neural network are to identify whether there is a basic geometric structure, so it can be transferred. The latter layers are often abstract and cannot be transferred. So, which layers to be transferred are case by case.
	
	In the work, we use front layers as feature extractors in the case of imitation learning. The decision model in the cloud can be used as the initial model for local training. In this way, cloud model can play a guiding role. It can speed up local training and increase the accuracy of local robots. In the training of local robots, the feature extraction layer is frozen and only the full connection layers are trained. If necessary, it can also adjust the relevant parameters in the process of back propagation. For example, some work may increase learning rate. After transfer learning, local robots successfully utilize knowledge from other robots in cloud robotic systems.
	\section{Experiments}
	In this section, we describe our experimental setup and present the results of FIL. To verify the effectiveness of FIL, we need to answer two questions: 1) Is FIL capable of generating an effective shared model based on shared knowledge in cloud robotic systems? 2) Does the shared model of FIL improve the learning process or accuracy of local robots with the transferring method? To answer the former question, we conduct experiments to generate cloud models and compare its performance with general models. To answer the latter question, we conduct experiments to compare the learning process and accuracy of local robots with FIL and without FIL.
	\subsection{Experimental setup}
	Self-driving car is regarded as an advanced robot. So there are sufficient reasons to use cars to verify robot control algorithms. In the work, we use Microsoft AirSim and CARLA as our simulator to evaluate the presented approach. In addition to having high-quality environments with realistic vehicle physics, AirSim and CARLA have a python API which allows for easy data collection and control. 
	\begin{figure}[thpb]
		\centering
		\subfigure[Data collection for local]{
			\label{fig:subfig:a} 
			\includegraphics[height=1.18in]{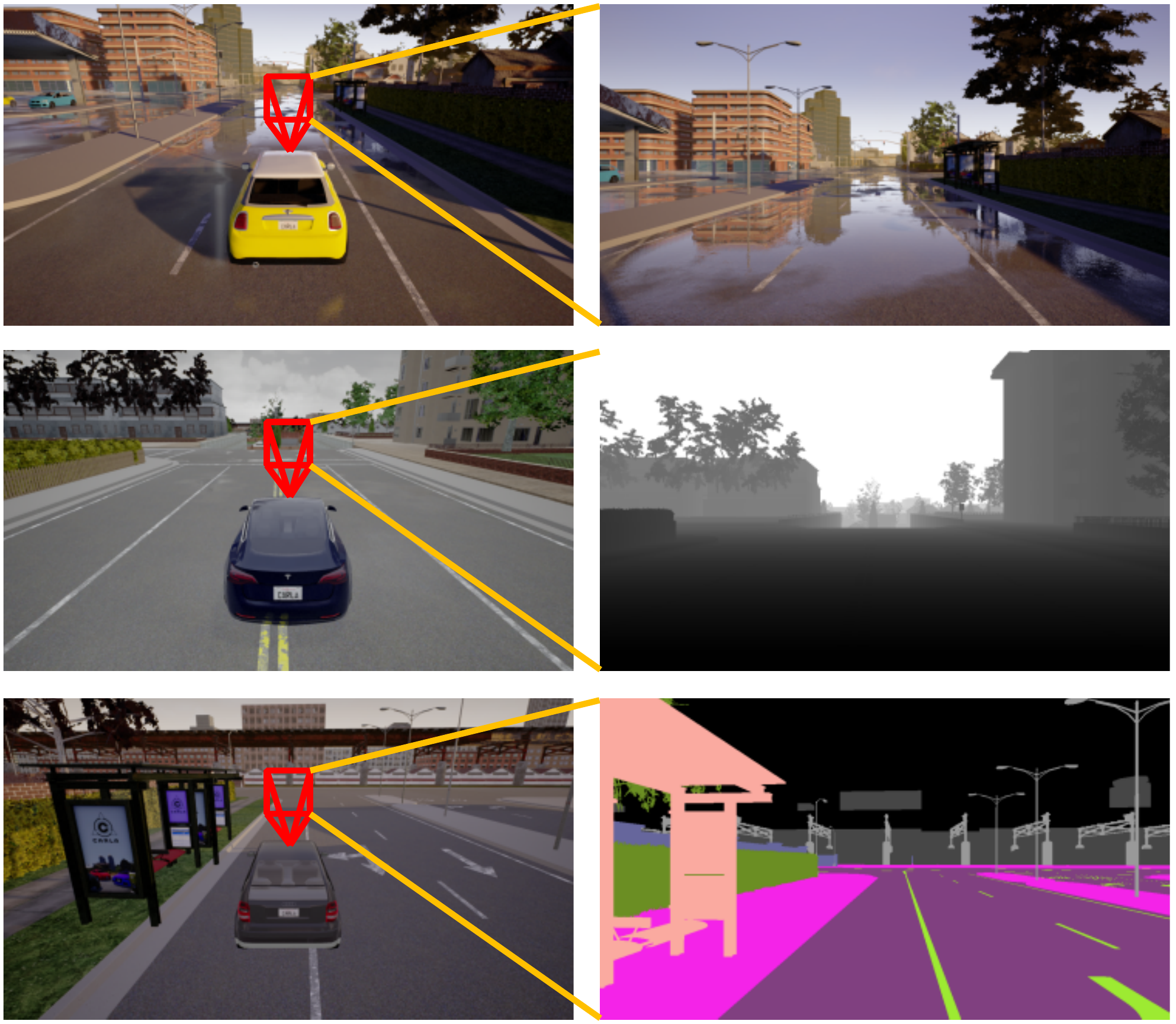}}
		\subfigure[Data collection for the cloud]{
			\label{fig:subfig:b} 
			\includegraphics[height=1.18in]{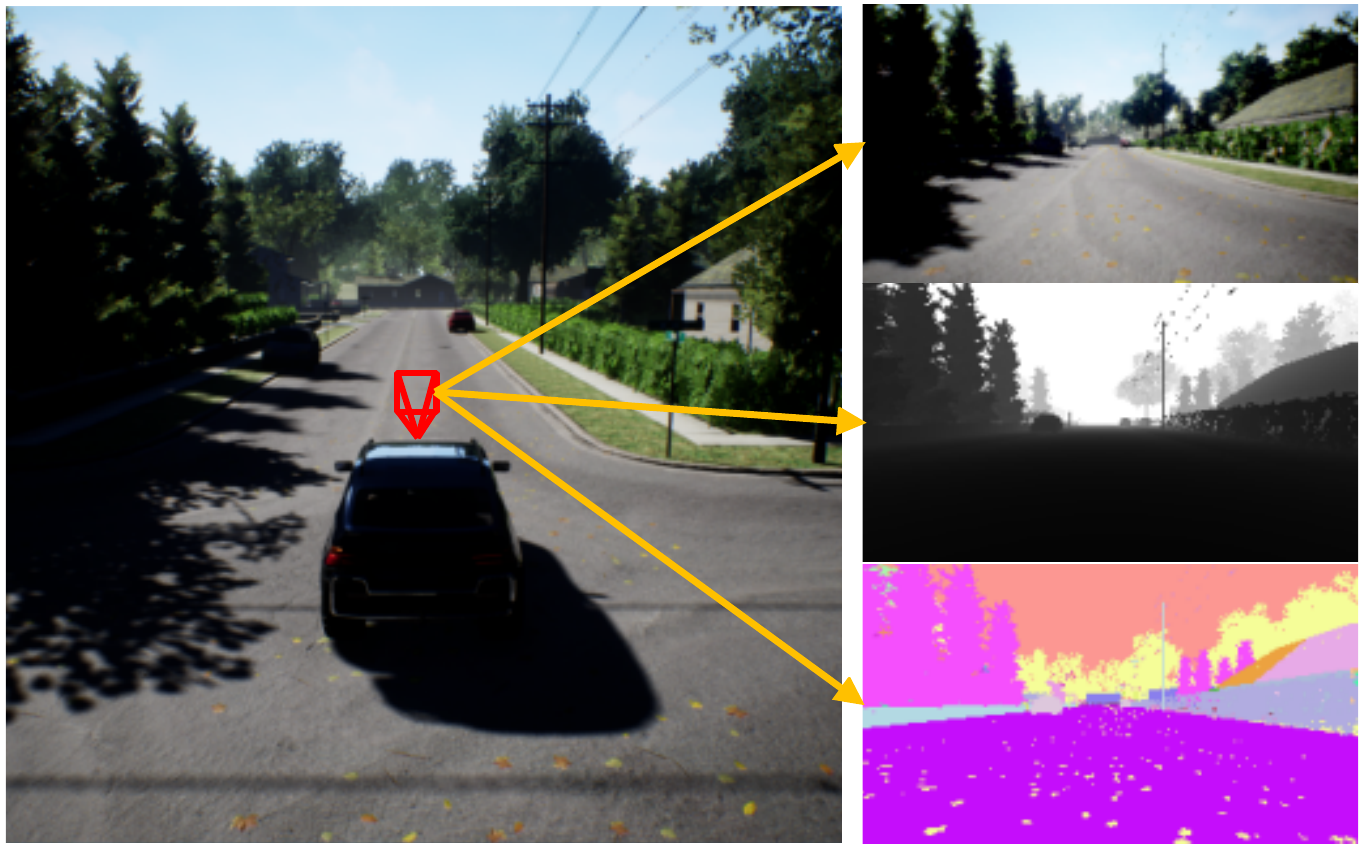}}
		\caption{As present in subfigure a, we collected training data for local robots in three different environments corresponding to three different types of data. As present in subfigure b, we collected different types but simultaneous data in many different environments.}
		\label{fig:subfig} 
	\end{figure}
	
	In order to collect training data, a human driver is presented with a first-person view of the environment (central camera). The driver controls the simulated vehicle using the keyboard. The driver keeps the car at a speed around 6m/s and strives to avoid collisions with cars or pedestrians, but ignores traffic lights and stop signs. As Fig.5 presents, we use three different types of sensor data: RGB images, depth images and semantic segmentation images. Because any of these three types of sensor data can make the agent to perform obstacle avoidance tasks within tolerable errors. The policy network mainly used convolution layers and fully connected layers. Then the output is used to produce the discrete action probabilities by a linear layer, followed by a softmax, and the value function by a linear layer. The architecture of the policy network similar to VGG-16 is presented in Fig. 5.
	\begin{figure}[thpb]
		\centering
		\includegraphics[width=0.6\linewidth]{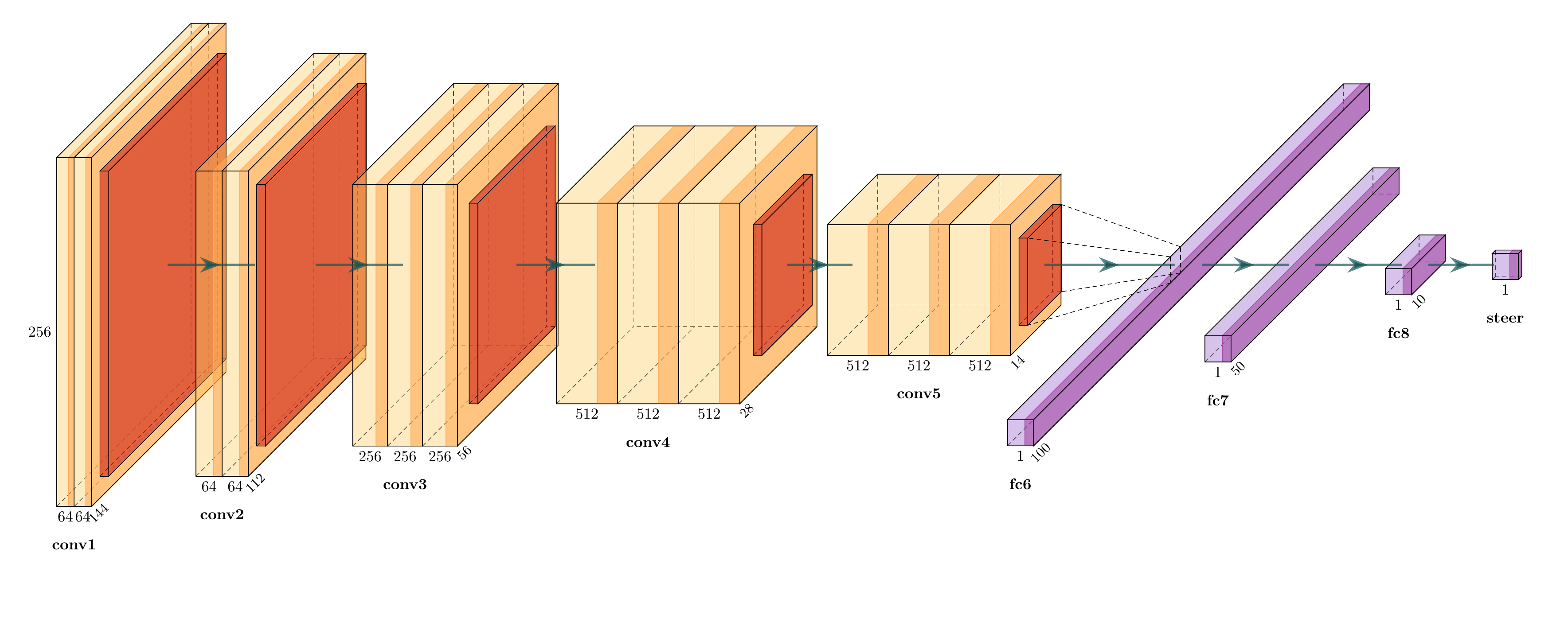}
		\caption{Network architecture of imitation learning in experiments}
		\label{fig:architecture}
	\end{figure}
	
	As for the cloud robotic systems, the server and agent communicate via HTTP requests, with the data server utilizing the Django framework to respond to asynchronous requests. Our cloud programs run on the Microsoft Azure cloud. We conduct local robot experiments with a single NVIDIA Quadro RTX 6000, which allows us to run our simulator to receive photo-realistic images for training. 
	\begin{figure*}[thpb]
		\centering
		\includegraphics[width=0.9\linewidth]{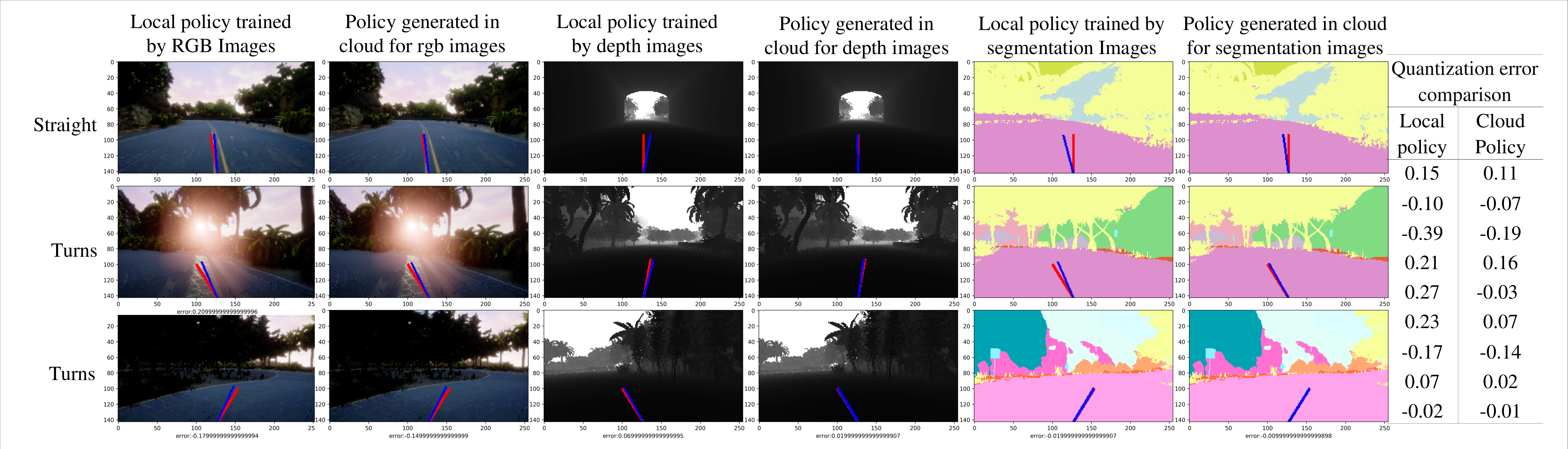}
		\caption{Performance comparison of local models and cloud generated models. The table on the right present quantization error of every image. Numbers from left to right correspond to figures from left to right.}
		\label{fig:architecture}
	\end{figure*}
	\subsection{Evaluation for the shared-model generating method in FIL}
	
	In this section, the work conducts an experiment to verify the effectiveness of the method for generating shared model in the cloud and present the performance of the shared model. The robot (car) should challenge the tasks such as avoiding collisions and making timely turns. The observations (images) are recorded by one central camera. The recorded control signal is the steering angle. The steering angle is scaled between -1 and 1, with extreme values corresponding to full left and full right, respectively. Considering the actual driving, we transfer the steering angle between -0.69 radians and 0.69 radians.
	
	After training the three private policy networks, the cloud will work to fuse their knowledge. As mentioned before, we assume that there are three companies train their policies by imitation learning with heterogeneous sensor data. We are not allowed to share the training data between these agents or send the raw data to the cloud. So the cloud server only gets the parameters of the three local networks and performs the knowledge fusion algorithm. Before that, we should collect different types of sensor data simultaneously from different environments in CARLA and Airsim. Then we will get the labels of every scene in cloud server. The cloud generates local policy network 1 for company 1 based on RGB images, local policy network 2 for company 2 based on semantic segmentation images, and local policy network 3 for company 3 based on infrared images. Then the parameters of these three private networks without raw training data are uploaded to the cloud server. Finally, the cloud server gets labels of shared scenes in the cloud. It is capable of generating policy networks corresponding to different sensor requests. In the experiment of this work, we used three types of scene data. So there were three policy networks generated. We named them cloud policies. The process in the cloud is unsupervised learning. There is no manual labels but cloud generation. However, in order to compare the performance between local policies and cloud policies, we labeled some scenes to mark the result. In the simulation environment, the controller uses the policy network to control the robot.
	\begin{table}[htbp]
		\centering
		\caption{Results of local policies and cloud policies.}
		\setlength{\tabcolsep}{1mm}{
			\begin{tabular}{cccc}
				\toprule
				Controller & \multicolumn{1}{p{3.3em}}{Hit the obstacle} & \multicolumn{1}{p{3.3em}}{Miss turns} & \multicolumn{1}{p{4.2em}}{Mistakes in straight} \\
				\midrule
				Local controller for RGB images & 3.45\% & 12\%  & 16.67\% \\
				Cloud controller for RGB images & 0.69\% & 0     & 0 \\
				Local controller for depth images & 0     & 20\%  & 0 \\
				Cloud controller for depth images & 0     & 4\%   & 0 \\
				Local controller for segmentation images & 0     & 12\%  & 6.67\% \\
				Cloud controller for segmentation images & 0     & 4\%   & 0 \\
				\bottomrule
		\end{tabular}}%
		\label{tab:addlabel}%
	\end{table}%
	
	Fig. 6 presents results of these six policy networks in some main challenging scenes to the car. From the Fig.6, we can see that compared with local models, the cloud models present higher accuracy. The cloud model acquired knowledge from different sensors and scenes. So that it can avoid errors of local models training from single training set collected by one type of sensors. We conducted 3 experiments, each one sets a different starting point. Then we evaluated the performance of robots in obstacle avoidance, turning and straight forward tasks. The results are summarized in Table 1. It can be seen from the experimental results that the cloud knowledge improves the local controller that is trained using general imitation learning. The controller based on cloud policies performs better. Especially the controller for RGB images.
	\subsection{Evaluation for the knowledge-transfer ability of FIL}
	\begin{figure}[thpb]
		\centering
		\includegraphics[width=0.8\linewidth]{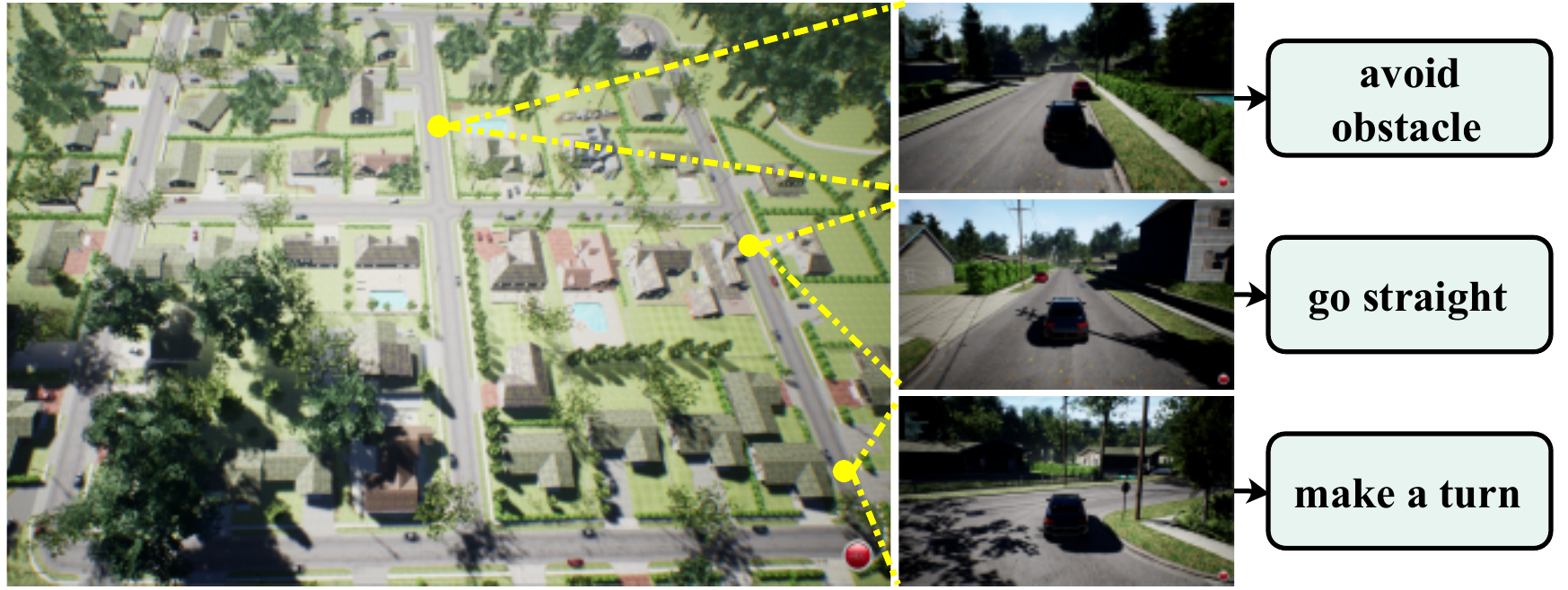}
		\caption{Challenges in the self-driving task}
		\label{fig:architecture}
	\end{figure}
	\begin{figure*}[thpb]
		\centering
		\includegraphics[width=0.8\linewidth]{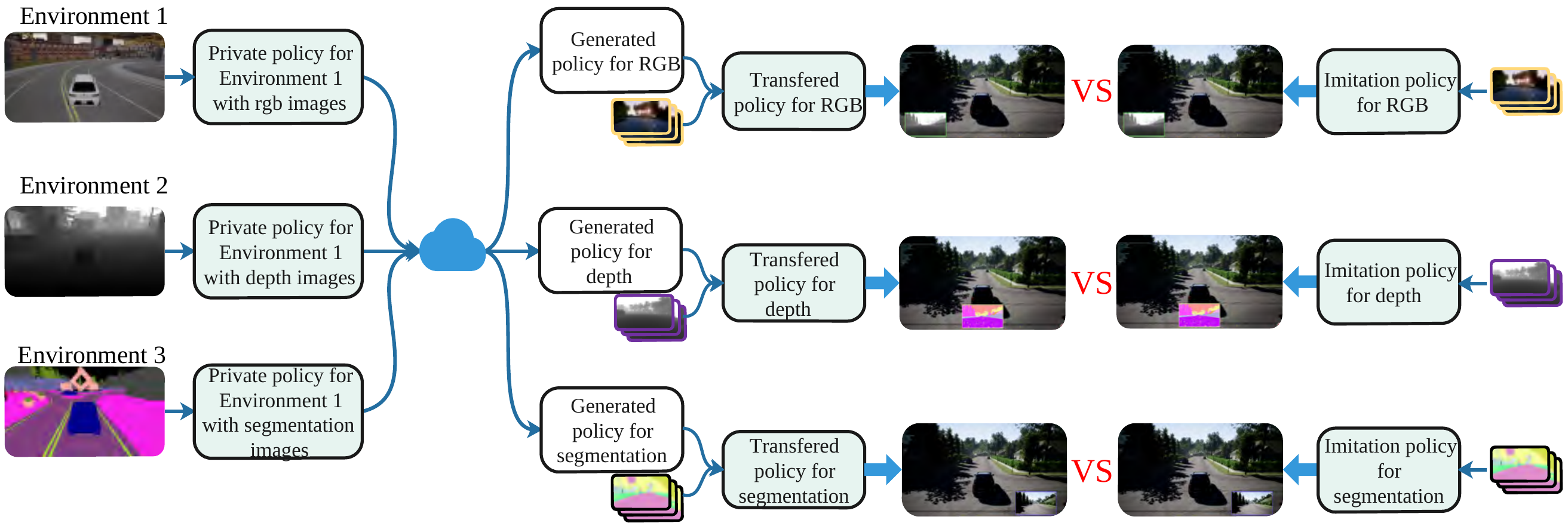}
		\caption{Procedure of the experiment for evaluating knowledge-transfer ability. Firstly, we performed imitation learning in three different environments with three types of sensor data (RGB images, depth images and semantic segmentation images) and got three policies. Secondly, the parameters of three private policies were sent to the cloud. The cloud generated labels of its own data base. Here is something to explained is that there are semantic segmentation difference between CARLA and Airsim, so there is a unifying process for semantic segmentation images. Thirdly, three policies corresponding to different sensor data were generated. After that, we performed transferring learning with corresponded type of images and obtained transferred policies. For comparison, we also conduct the general imitation learning experiment on the right part of the image. Finally, we compare these policies in a new environment (neighborhood in Airsim). }
		\label{fig:architecture}
	\end{figure*}
	\begin{figure*}[thpb]
		\centering
		\includegraphics[width=0.9\linewidth]{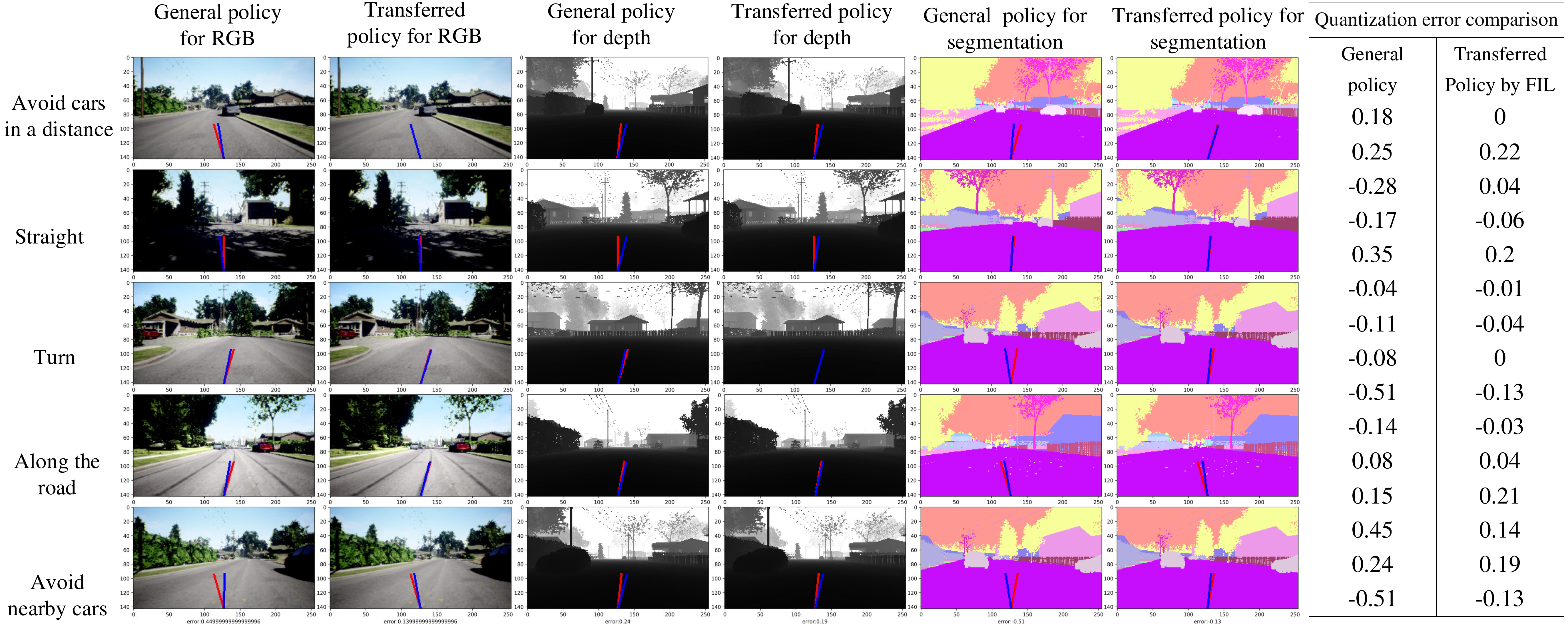}
		\caption{Performance comparison of general policies and transferred policies. The table on the right present quantization error of every image. Numbers from left to right correspond to figures from left to right.}
		\label{fig:architecture}
	\end{figure*}
	\begin{table*}[htbp]
		\centering
		\caption{}
		\begin{tabular}{cccccc}
			\toprule
			Controller & Error rate in normal & Error rate in rain & Error rate in snow & Error rate in fog & Error rate in dust \\
			\midrule
			General controller for RGB images & 17.39\% & 26.09\% & 30.43\% & 34.78\% & 52.17\% \\
			Transferred controller for RGB images & 4.35\% & 8.70\% & 17.39\% & 31.82\% & 39.13\% \\
			General controller for depth images & 13.04\% & 4.35\% & 8.70\% & 8.70\% & 17.39\% \\
			Transferred controller for depth images & 10.87\% & 4.35\% & 8.70\% & 8.70\% & 15.22\% \\
			General controller for segmentation images & 2.17\% & 2.17\% & 6.52\% & 21.74\% & 36.36\% \\
			Transferred controller for segmentation images & 2.17\% & 2.17\% & 5.43\% & 17.39\% & 31.82\% \\
			\bottomrule
		\end{tabular}%
		\label{tab:addlabel}%
	\end{table*}%
	We conducted the experiment as illustrated in Fig. 7 to evaluate the knowledge-transfer ability in FIL. As presented in Fig. 8, we compare the six models in a neighborhood environment. Corresponding to every type of training data, we obtained a pair of policies: a transferred policy and a general policy. So there are three pairs of policies generated in the experiment. The performance of controllers based on these policies in key challenging tasks are presented in Fig. 9. The results are summarized in Table 2. From the results, we can see that the imitation learning models obtained in cloud robotic system perform significantly better in accuracy, compared with general models that trained by traditional imitation learning without shared knowledge. 
	FIL improves the training process of imitation learning with the help of shared knowledge. There is a pre-trained model from the cloud for transfer in local imitation learning. So there is no need for local robots to learn from scratch. We present the comparison of train process in Fig 10. From the figure, we can see that the transferred policies have lower error starting point and the error value.
	\begin{figure}[thpb]
		\centering
		\includegraphics[width=1\linewidth]{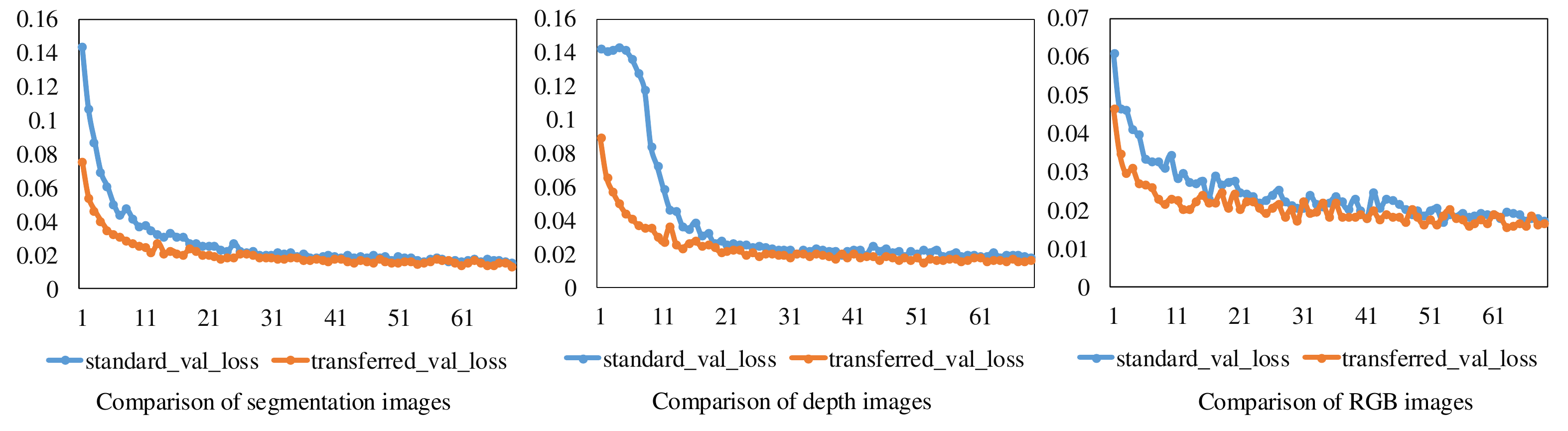}
		\caption{Training process comparison of general learning and transfer learning in FIL}
		\label{fig:architecture}
	\end{figure}
	\begin{figure}[thpb]
		\centering
		\subfigure[Rain]{
			\label{fig:subfig:a} 
			\includegraphics[width=0.77in]{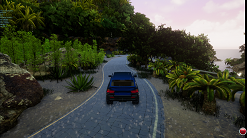}}
		\subfigure[Snow]{
			\label{fig:subfig:b} 
			\includegraphics[width=0.77in]{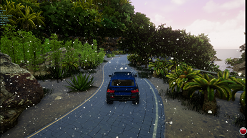}}
		\subfigure[Fog]{
			\label{fig:subfig:a} 
			\includegraphics[width=0.77in]{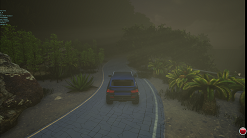}}
		\subfigure[Dust]{
			\label{fig:subfig:b} 
			\includegraphics[width=0.77in]{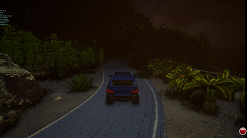}}
		\caption{Bad weather conditions, rain, snow, fog, sand and dust.We conducted data collection and comparison experiments in these environments}
		\label{fig:subfig} 
	\end{figure}
	Local policy models transferred by FIL also have better generalization. Controllers based on transferred policies from FIL perform better in different weather compared with policies trained by general imitation learning. As presented in Fig.11, we conducted the experiments for controllers in different weathers. The results are presented in the last three rows of the Table 2. The results showed that the model from FIL could improve the accuracy of the controller from general imitation learning in bad weather.
	\section{conclusion}
	This work proposes a privacy considered imitation learning framework for cloud robotic systems with heterogeneous sensor data, named Federated Imitation Learning (FIL). FIL is capable of improving imitation learning efficiency and accuracy of local robots by taking advantage of knowledge from other robots in the cloud robotic system. Additionally, we propose the knowledge fusion algorithm and introduce a transfer methods in FIL. Our approach is able to fuse heterogeneous knowledge and protect privacy of local robots. Last but not least, we validated our framework and algorithms in policy-learning experiments.
	
	The proposed framework FIL only improves the imitation learning approach in cloud robotic systems. We leave it as future work to make FIL flexible to deal with different learning approaches of local robots. Various types of sensor data are utilized in robotic tasks, thus extensible fusion method for more types of data is also a worthy research issue in the future. A more flexible FIL will provide a wider range of services in cloud robotic systems.

\end{document}